\documentclass[lettersize,journal]{IEEEtran}
\usepackage{amsmath,amsfonts}
\usepackage{algorithmic}
\usepackage{array}
\usepackage[caption=false,font=scriptsize,labelfont=rm,textfont=rm]{subfig}
\usepackage{textcomp}
\usepackage{stfloats}
\usepackage{url}
\usepackage{verbatim}
\usepackage{float}
\usepackage{graphicx}
\usepackage{cite}
\usepackage{hyperref}
\usepackage{amssymb}
\usepackage{pifont}
\usepackage{enumerate}
\usepackage{graphicx}
\usepackage{multirow}
\usepackage{amsmath}
\usepackage{booktabs}
\usepackage{xcolor}
\definecolor{hc}{RGB}{0,0,255}

\makeatletter

\usepackage{bm}
\usepackage[ruled,vlined,linesnumbered]{algorithm2e}
\SetKwInOut{Input}{Input}
\SetKwInOut{Output}{Output}

\usepackage{placeins}
\begin{document}
	
	\title{SE-GNN: Seed Expanded-Aware Graph Neural Network with Iterative Optimization for Semi-supervised Entity Alignment}
	
		\author{IEEE Publication Technology,~\IEEEmembership{Staff,~IEEE,}
		\author{Tao~Meng, Shuo~Shan, Hongen Shao, Yuntao Shou, Wei~Ai, and~Keqin~Li,~\IEEEmembership{Fellow,~IEEE}
			\thanks{Corresponding Author: Tao Meng~(mengtao@hnan.edu.cn)}
			
			\IEEEcompsocitemizethanks{
				\IEEEcompsocthanksitem T. Meng, S. Shan, H. Shao, Y. Shou and W. Ai are with School of Computer and Information Engineering, Central South University of Forestry and Technology, Hunan, Changsha 410004, China. (aiwei@hnan.edu.cn, shanshuo@csuft.edu.cn, mengtao@hnan.edu.cn)
				\IEEEcompsocthanksitem K. L is with the Department of Computer Science, State University of New York, New Paltz, New York 12561, USA. (lik@newpaltz.edu)}}
		\thanks{This paper was produced by the IEEE Publication Technology Group. They are in Piscataway, NJ.}
		\thanks{Manuscript received April 19, 2021; revised August 16, 2021.}}
	
	\IEEEpubid{}
	
	\maketitle
	
	\begin{abstract}
		
		Entity alignment aims to use pre-aligned seed pairs to find other equivalent entities from different knowledge graphs (KGs) and is widely used in graph fusion-related fields. However, as the scale of KGs increases, manually annotating pre-aligned seed pairs becomes difficult. Existing research utilizes entity embeddings obtained by aggregating single structural information to identify potential seed pairs, thus reducing the reliance on pre-aligned seed pairs. However, due to the structural heterogeneity of KGs, the quality of potential seed pairs obtained using only a single structural information is not ideal. In addition, although existing research improves the quality of potential seed pairs through semi-supervised iteration, they underestimate the impact of embedding distortion produced by noisy seed pairs on the alignment effect. In order to solve the above problems, we propose a seed expanded-aware graph neural network with iterative optimization for semi-supervised entity alignment, named SE-GNN. First, we utilize the semantic attributes and structural features of entities, combined with a conditional filtering mechanism, to obtain high-quality initial potential seed pairs. Next, we designed a local and global awareness mechanism. It introduces initial potential seed pairs and combines local and global information to obtain a more comprehensive entity embedding representation, which alleviates the impact of KGs structural heterogeneity and lays the foundation for the optimization of initial potential seed pairs. Then, we designed the threshold nearest neighbor embedding correction strategy. It combines the similarity threshold and the bidirectional nearest neighbor method as a filtering mechanism to select iterative potential seed pairs and also uses an embedding correction strategy to eliminate the embedding distortion. Finally, we will reach the optimized potential seeds after iterative rounds to input local and global sensing mechanisms, obtain the final entity embedding, and perform entity alignment. Experimental results on public datasets demonstrate the excellent performance of our SE-GNN, showcasing the effectiveness of the model. Our code is publicly available at https://github.com/taomeng/SE-GNN. 
	\end{abstract}

	\section{Introduction}
	\label{section: section1}
\begin{figure}
		\centering
		\includegraphics[width=1\linewidth]{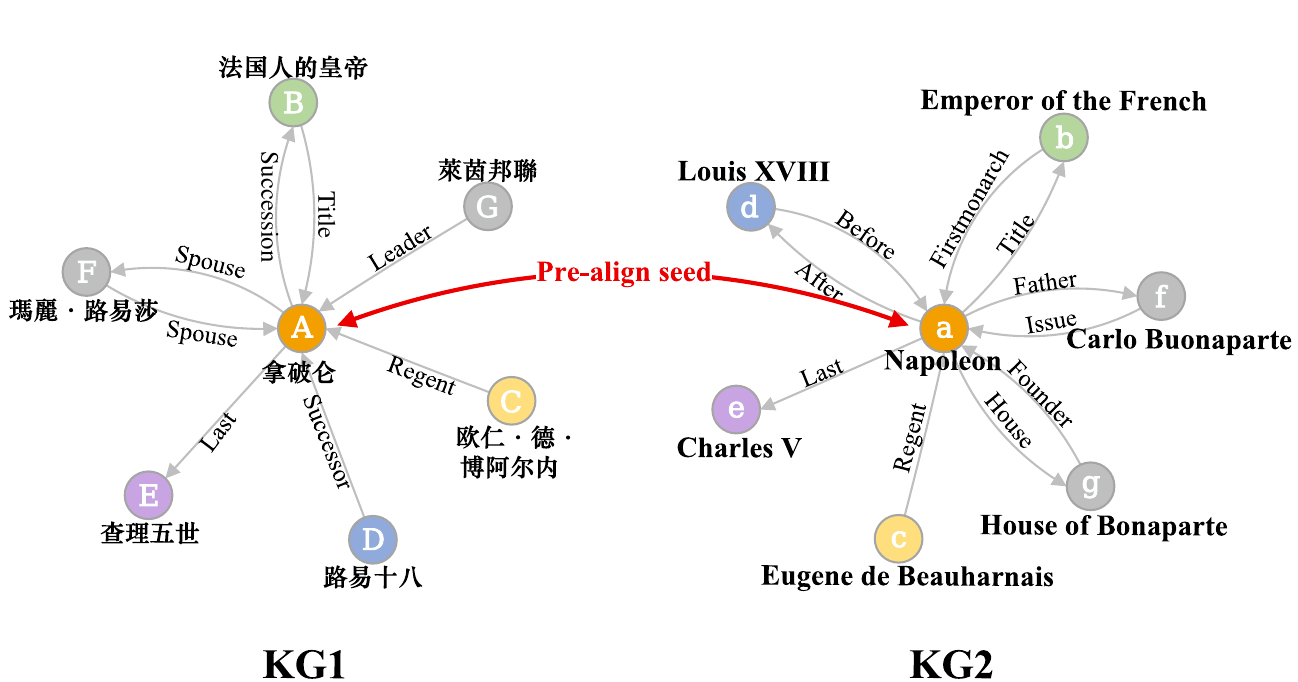}
		\caption{A cross-language knowledge graph composed of different data sources. Among entities with colored backgrounds, entities of the same color are equivalent. The entities in the gray background are non-equivalent entities. Both KGs have incomplete information about Napoleon. }
		\label{fig1}
	\end{figure}

	\IEEEPARstart
	{K}{n}owledge graphs (KGs) is a knowledge representation method of graph structure used to describe entities and their relationships. Common KGs include DBpedia\cite{auer2007dbpedia}, YAGO\cite{suchanek2007yago} and Freebase\cite{bollacker2008freebase}, which play an important role in research fields such as information extraction\cite{distiawan2019neural}, recommendation systems\cite{ko2022survey}, graph question answering\cite{soares2020literature}. However, KGs are composed of multiple heterogeneous data sources, their coverage is limited, and they suffer from incomplete entity descriptions. As shown in Fig.\ref{fig1}, KG1 lacks information on “House of Bonaparte” and “Carlo Buonaparte” while KG2 lacks details on “Confederation of the Rhine” and “Marie Louisa”. Neither KG provides a comprehensive description of “Napoleon”. In order to expand the coverage and knowledge areas of KGs, it is necessary to integrate information between different graphs through KG fusion\cite{zhao2020multi} to describe entities more comprehensively. Entity alignment\cite{zhao2020experimental} is a key step in knowledge graph fusion. It can merge information between knowledge graphs and provide richer and more accurate entity descriptions. 
	
	Entity alignment uses pre-aligned seed pairs as a bridge to achieve the matching and linking of equivalent entities in KGs. Early entity alignment was primarily based on the idea of TransE\cite{bordes2013translating}. They\cite{lin2015learning,zhu2019neighborhood,yang2020cotsae} viewed the “relation” between entities in the KGs as a “translation” of the head entity and tail entity in the vector space, making equivalent entities closer in the vector space. However, these methods cannot fully utilize the graph structure information in the KGs, resulting in the inability to effectively mine the complex correlations between entities. 
	In recent years, GNNs\cite{kipf2016semi,zou2022multi, shou2025masked, shou2023comprehensive, meng2023deep, meng2024multi, meng2024deep, shou2024low, ai2024edge, shou2024adversarial, ai2024gcn, ai2023gcn} has made up for this shortcoming of the TransE model with its excellent graph data modeling capabilities and therefore is widely used for entity alignment. GNN-based entity alignment methods\cite{wang2018cross,fang2023relation,zhu2023effective, ai2023two, shou2023graph, shou2023czl, shou2024contrastive} aggregate structural, relational, and attribute information of entities and their neighbors to generate rich and context-aware embeddings, effectively enhancing the semantic representation capability of entities. In addition, considering the important role of pre-aligned seed pairs in entity alignment tasks, some methods also use semi-supervised iteration strategies\cite{sun2018bootstrapping,zhu2023semi,zhang2023semi, shou2024revisiting, meng2024masked, shou2024efficient, meng2024revisiting} to construct potential seed pairs to expand the seed set and thereby improve the alignment effect. While these methods have made progress in entity alignment, there are still two pressing issues that need to be addressed.

	{\textbf{(1) Heavy reliance on single structural information to expand seed.}}
	Existing entity alignment methods mainly utilize pre-labeled equivalent entity pairs as pre-aligned seed pairs. Theoretically\cite{mao2020mraea,sun2018bootstrapping, shou2024spegcl, shou2024graph}, the more seed pairs there are, the more comprehensive information the model can collect and the better the alignment effect. However, with the expansion of knowledge graphs, manually labeling seed pairs becomes time-consuming and expensive, which limits the widespread application of entity alignment technology. To overcome this limitation, some methods\cite{zhang2023semi,cai2022semi, shou2024dynamic, zhang2024groupface, ying2021prediction, shou2025contrastive, zhang2024multi} use entity embeddings obtained by iteratively aggregating single structural information to identify potential seed pairs, thereby expanding the seed set and enhancing model performance. Although these methods reduce the reliance on manually labeled seed pairs, due to the structural heterogeneity of KGs\cite{bing2023heterogeneous, shou2025dynamic, ai2024mcsff, ai2024graph, fu2025sdr, ai2024seg}, relying solely on single structural information may limit the mining of more and higher-quality potential seed pairs. Furthermore, these methods often ignore the critical role of entity semantic attributes and information fusion in the seed screening process. Therefore, how to effectively combine the structural features and semantic attributes of entities to optimize the seed pair screening process has become an academic challenge that needs to be solved urgently.

	{\textbf{(2) Underestimating embedding distortion caused by noise seeds.}} In entity alignment tasks, the loss function is usually constructed by minimizing the distance of positive samples and maximizing the distance of negative samples\cite{zhao2020experimental,sun2018bootstrapping, zhang2025lra, shou2023graphunet, shou2022object}. When semi-supervised entity alignment selects potential seed pairs through a screening strategy, adulterated noisy seed pairs will be incorrectly labeled as positive samples. Due to the interference of this erroneous information, the embedding distance between noise seed pairs will continue to shrink. This embedding distortion phenomenon causes the position of the noise seed pair to be incorrectly shifted in the vector space, affecting the similarity of the actual positive samples in the embedding representation and the correlation of the related entity pairs in the embedding representation. 
	Existing studies \cite{mao2020mraea,zhu2023semi,whang2023data} have attempted to reduce the impact of noise by setting a high confidence threshold, implementing a two-way election strategy, or optimizing the loss function. Although certain effects have been achieved, considering the inherent complexity of EA, including factors such as data incompleteness, inconsistency, and noise diversity, effectively distinguishing between accurate signals and noise and eliminating the impact of noise still faces considerable difficulties. In addition, the embedding distortion caused by the noise seed pairs further exacerbates the deviation of the alignment results and affects the overall accuracy. Thus, how to effectively alleviate the embedding distortion problem while reducing noise seed pairs is also a problem that needs to be overcome in current research.

	In order to address the challenges above, this paper proposes a novel seed expanded-aware graph neural network with iterative optimization for semi-supervised entity alignment, named SE-GNN. To commence, SE-GNN proposes a seed expansion strategy based on neighborhood-level semantic information. It comprehensively utilizes entities' semantic attributes and structural features, combined with a conditional filtering mechanism, to obtain high-quality initial potential seed pairs and enhance the richness of alignment information. Subsequently, SE-GNN designs local and global awareness mechanisms. While introducing initial potential seed pairs, it also profoundly mines local and global information in KG to obtain a more comprehensive entity embedding representation, alleviates the impact of KG structural heterogeneity, and lays a solid foundation for subsequent optimization of initial potential seed pairs. Next, SE-GNN constructs a threshold nearest neighbor embedding correction strategy during iterative optimization. This strategy combines similarity threshold and bidirectional nearest neighbor method to select iterative potential seed pairs. Meanwhile, it uses Xavier initialization to modify entity embeddings, thereby eliminating the embedding distortion caused by noise seeds and mitigating the risk of gradient vanishing in the model. Finally, SE-GNN enters the iterative potential seed pairs that satisfy the optimization round into the local and global awareness mechanisms to obtain the final entity embedding and perform entity alignment. The main contributions of this paper are summarized as follows:

	\begin{itemize}
		
	\item We design an innovative method that comprehensively utilizes the semantic attributes and structural features to obtain potential seed pairs. This method can fully capture the alignment signals within the KG, reduce the dependence on single structural information, and improve the robustness of seed pair recognition.
	\item We design a novel local and global awareness mechanism. It combines the relation and entity semantic information in KG to generate entity embeddings with local and global features, alleviating the impact of KG's structural heterogeneity on alignment performance. 
	\item We propose an efficient threshold nearest neighbor embedding correction strategy. While selecting iterative potential seed pairs, it also uses an embedding correction method to eliminate the embedding distortion caused by noisy seed pairs.
	\item We have conducted many experiments on different data sets, and SE-GNN has achieved excellent results. It thoroughly verifies the remarkable effectiveness of SE-GNN in alleviating the dependence on structural information and alleviating embedding distortion. 
	\end{itemize}
	
	The remaining part of the article describes the following aspects. Section \ref{section: section2} reviews the entity alignment preliminaries and related work, and Section \ref{section: section3} introduces the model we proposed in detail. Section \ref{section: section4} discusses experimental settings and experimental results. Section \ref{section: section5} summarized our work and discussed the outlook for future work. 
	
	\section{RELATED WORK}
	\label{section: section2}
	
	This section reviews the techniques involved in entity alignment in the paper. Three methods are included to solve the entity alignment task: the translation-based entity alignment method, the GNN-based entity alignment method, and the semi-supervised entity alignment method.

	\subsection{\textbf{Translation-Based Entity Alignment Methods}}
	Entity alignment based on representation learning requires mapping entity vectors from different knowledge graphs into a unified vector space to calculate the similarity and distance between entities. TransE\cite{bordes2013translating} has received widespread attention from researchers because of its excellent performance in repressentation learning. It treats the tail entity of a triple as adding the head entity and relation in the vector space, continuously adjusting the values of the head entity, relation, and tail entity to satisfy the condition $h+r \approx t$ as much as possible. Building upon this concept, researchers have proposed entity alignment methods based on translation. 
	
	MTransE\cite{chen2017multilingual} firstly adopts TransE to address entity alignment tasks. It encodes entities and relations from knowledge graphs in different languages into independent spaces, computes entity embeddings in these spaces, and maps transformations to other independent spaces, thereby achieving entity alignment across multilingual knowledge graphs. TransEdge\cite{sun2019transedge} introduces context information of relationships to differentiate the same relationships in different entities, enhancing the TransE model's capability in handling complex relationships. However, the above methods only consider information based on triples, neglecting using other information in knowledge graphs. 
	
	Therefore, JAPE\cite{sun2017cross} builds on the entity representation obtained using the TransE model and obtains attribute representations using Skipgram, thereby combining structural information and attribute information to obtain more entity semantic information. In addition, COTSAE\cite{yang2020cotsae} derives entity attribute information from attribute types and values and learns the attention distribution of attribute types and attribute values through joint attention. The above methods utilize relationship, neighborhood, or attribute information but mainly focus on local information based on triples, unable to consider entity alignment from the graph's neighborhood structure. Some GNN-based methods have been proposed to address this limitation. 
	
	\subsection{\textbf{GNN-Based Entity Alignment Methods}}
	Compared to traditional neural networks, GNNs can effectively capture complex relationships and contextual information in knowledge graphs, thus better modeling and understanding graph data. It enables GNN-based models to demonstrate outstanding capabilities in entity alignment. GCN-Align\cite{wang2018cross} for the first time applies GCN\cite{kipf2016semi} to entity alignment tasks, achieving excellent results by aggregating semantic and attribute information of entities through GCN. Subsequently, NMN\cite{wu2020neighborhood} combines GCN modeling with neighborhood sampling methods to capture rich neighborhood features. ERGCN\cite{fang2023relation} learns entity and relationship embeddings simultaneously through entity convolution and relationship convolution and models relationship information through quadruples to obtain rich neighborhood information. Additionally, RHGN\cite{liu2023rhgn} distinguishes relations and entities in KG through relation gate convolution and solves neighbor heterogeneity and relation heterogeneity issues using cross-graph embedding exchange and soft relationship alignment. DMFNet\cite{guo2023variational} successfully combines multi-view similarity information to infer potential associations and adaptively extracts multi-level contextual embeddings. EAMI\cite{zhu2023effective} leverages GCN and Highway to model various information, obtaining more precise entity representations. 
	
	The GCN model fails to consider the important differences among neighboring nodes when aggregating neighbor node information. In contrast, GAT\cite{velivckovic2017graph} utilizes a self-attention mechanism\cite{vaswani2017attention} to assign different weights to different neighbor nodes, effectively addressing the issue above, hence widely applied in entity alignment. For instance, AliNet\cite{sun2020knowledge} combines gate strategies with GAT, enabling targeted weighted aggregation of neighborhood information when aggregating multi-hop neighborhoods. DVGNET\cite{li2023dual} starts from entities and relationships, assigns weights to neighbor nodes through GAT, and calculates the relationship matching degree based on relationship embedding, thereby alleviating the heterogeneity problem. Moreover, Dual-AMN\cite{mao2021boosting} achieves a fusion of inter-graph information through proxy matching vectors, significantly reducing the model's computational complexity. CTEA\cite{xu2024position} designs a joint embedding model that combines entity embedding, relationship embedding, and attribute embedding to generate transferable entity embedding. ASGEA\cite{luo2024asgea} constructs an Align-Subgraph using anchor links and designs a path-based graph neural network to identify and integrate logical rules across knowledge graphs. RoadEA\cite{sun2022revisiting} constructs attribute encoders and relationship encoders using attention mechanisms to learn entity embeddings and adopts an adaptive embedding fusion gate mechanism to integrate the two types of encoders. Although the above GNN-based methods have shown excellent results in entity alignment, they ignore the consideration of combining the semantic attributes on the graph to obtain high-order semantic neighbors. This limits their ability to aggregate the global information of entities.

	\begin{figure*}
		\centering
		\includegraphics[width=1\linewidth]{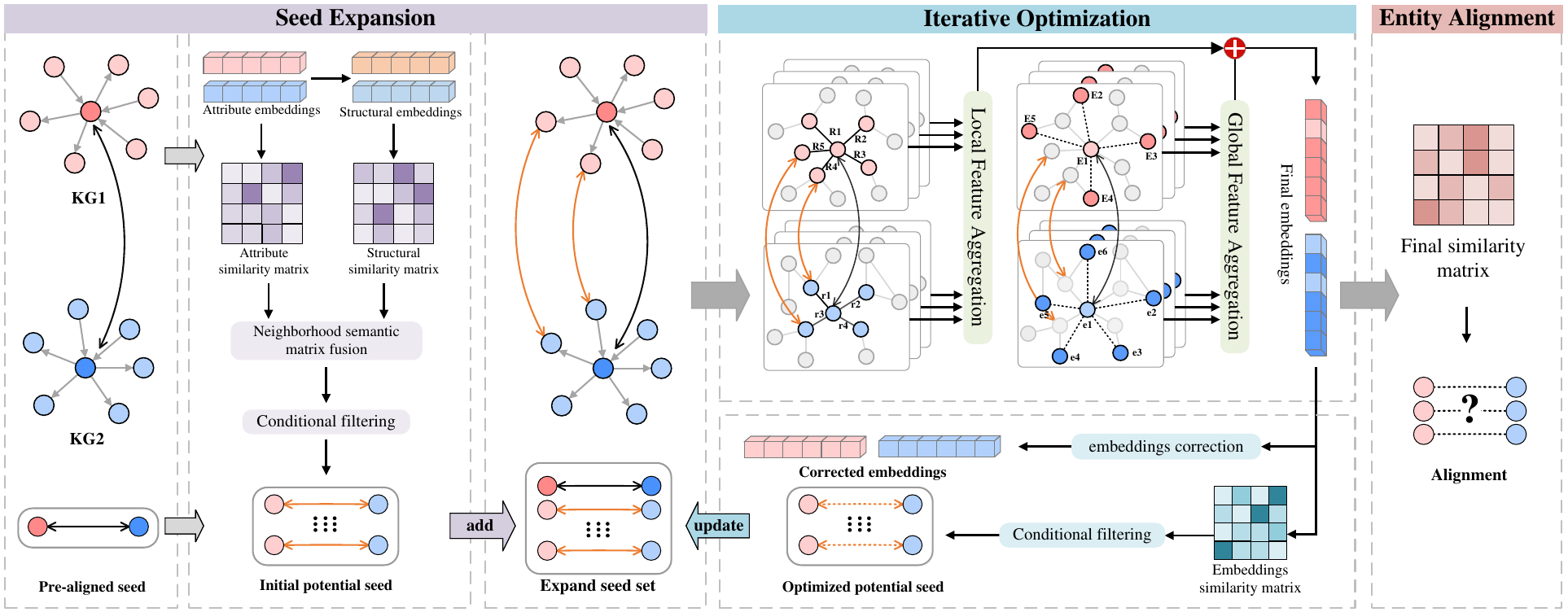}
		\caption{Framework diagram of SE-GNN. It consists of three parts: seed expansion, iterative optimization, and entity alignment. First, the seed expansion part obtains the initial potential seed through neighborhood-level semantic information and inputs it and the pre-aligned seed pair into the iterative optimization part. Next, the iterative optimization part optimizes seed pairs and corrects entity embeddings through local and global awareness mechanisms and threshold nearest neighbor embedding correction strategy. Finally, we will input the optimized potential seed pairs after the iteration round into the local and global awareness mechanism again to obtain the final entity embedding and perform entity alignment.}
		\label{fig2}
	\end{figure*}
	
	\subsection{\textbf{Semi-Supervised Entity Alignment Methods}}
	Entity alignment requires a seed set as training data. A rich seed set implies better results. However, manually labeling training seed pairs in large knowledge graphs with billions of entities is labor-intensive and inefficient. Therefore, researchers attempt to select potential seed pairs for alignment from unlabeled entities, iteratively expanding the seed set to improve alignment effects through semi-supervised training. IPTransE\cite{zhu2017iterative} and BootEA\cite{sun2018bootstrapping} proposed expanding the seed set earlier. IPTransE jointly learns entity and relation embeddings, calculates distances between entities in two ways, and adds highly confident entity pairs to the seed set. BootEA adopts alignment correction methods to remove potentially mislabeled seed pairs in subsequent training to reduce error accumulation during iteration.  
	
	The subsequent research will focus on improving the quality of potential seed pairs. MRAEA\cite{mao2020mraea} adopts a bidirectional iterative strategy, considering entity pairs that are mutual nearest neighbors as potential seed pairs. CUEA\cite{zhao2022toward} devises a mismatched entity prediction module to filter out incorrect seed pairs. GALA\cite{zhang2023semi} utilizes a custom confidence mechanism to expand the seed set. EASY\cite{ge2021make} adopts a structure-based refinement strategy to correct misaligned entities generated during training iteratively. RANM\cite{cai2022semi} adds bidirectional nearest neighbor entity pairs to the candidate set, and only those that remain nearest neighbors consecutively multiple times are added to the seed set. SNGA\cite{zhu2023semi} uses a bidirectional nearest neighbor iteration strategy to expand the seed set and further enhances the alignment effect through global matching. UPLR\cite{li2022uncertainty} adaptively mines trustworthy samples, enhancing domain similarity gradually to reduce the impact of noisy labels on entity embedding representations. 
	
	The above methods obtain potential seed pairs by calculating the embedding distance between entities and setting filter conditions. However, they do not consider using various information to construct initial potential seed pairs and ignore the embedding distortion caused by noise seed pairs on entity embedding. Compared with their models, SE-GNN combines semantic attributes and structural features to select initial potential seed pairs and adds an entity embedding correction method to eliminate embedding distortion.

	\section{PROPOSED METHOD}
	\label{section: section3}
	In this section, we introduce the task definition of entity alignment and then describe our semi-supervised entity alignment model, SE-GNN.

	\subsection{\textbf{Problem Description}}

	The knowledge graph is a database that describes entities and the relationships between entities. It is usually expressed as $G=(E,R,T)$, where $E$ represents the entity set, $R$ represents the relationship set, and $T$ represents the set of triples $\{(e_1,r,e_2) \mid e_1,e_2 \in E, r \in R\}$. In the entity alignment task, the source knowledge graph is usually represented as $G_1=(E_1, R_1, T_1)$, the target knowledge graph is represented as $G_2=(E_2, R_2, T_2)$, and the pre-aligned seed set is represented as $S=\{(u,v) \mid u \in E_1, v \in E_2, u \equiv v\}$. 
	
	Given two different knowledge graphs, $G_1$ and $G_2$, the goal of entity alignment is to match equivalent entities with the same meaning from these two knowledge graphs, combined with the pre-aligned seed set $S$, to achieve information fusion between knowledge graphs. 
	
	\subsection{\textbf{Overview of SE-GNN}}
	
	As shown in Fig.\ref{fig2}, our model consists of three parts: seed expansion, iterative optimization, and entity alignment. First, we calculate the neighborhood-level semantic information similarity between entities and then combine the similarity threshold with bidirectional nearest neighbors to select initial potential seed pairs for seed expansion. Next, we implement iterative optimization of seeds through local and global awareness mechanisms and threshold nearest neighbor embedding correction strategy. The local and global awareness mechanism mines local and global information to obtain a more comprehensive entity embedding representation. The threshold nearest neighbor embedding correction strategy combines similarity threshold and bidirectional nearest neighbor method to select iterative potential seed pairs. It also uses Xavier initialization to correct the entity embedding and eliminate the embedding distortion caused by noise seeds. Finally, we enter the iterative potential seed pairs that satisfy the optimization round into the local and global awareness mechanisms to obtain the final similarity matrix and perform entity alignment. 
	
	\begin{figure*}
		\centering
		\includegraphics[width=1\linewidth]{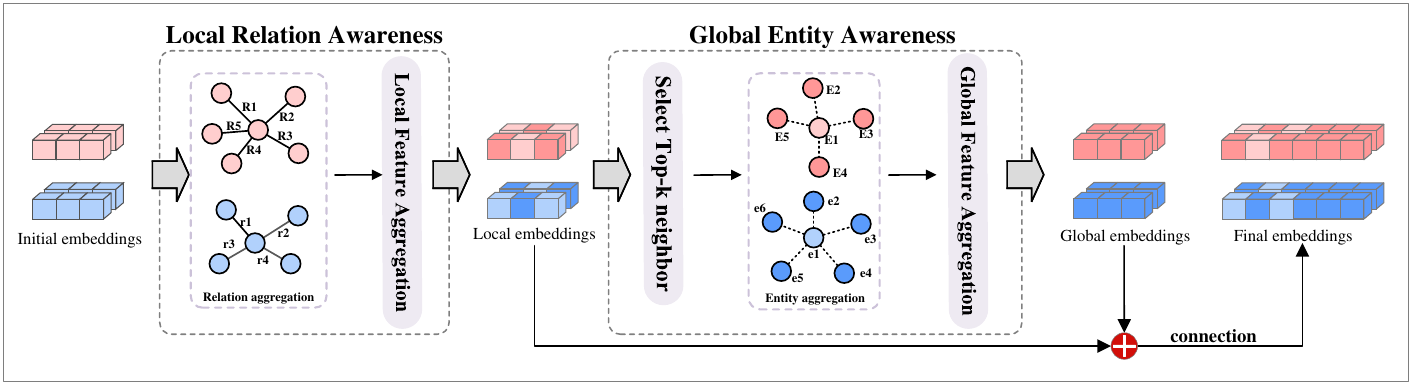}
		\caption{The details of the local and global awareness mechanism process include the local relation awareness module and global entity awareness module. }
		\label{fig3}
	\end{figure*}
	
	\subsection{\textbf{Seed Expansion}}
	
	In entity alignment, equivalent entities usually have similar neighbors. This structural similarity is also reflected at the entity semantic level. The neighbors of equivalent entities also exhibit a certain degree of similarity in their semantics. Therefore, we combine the semantic attributes and structural features of entities to obtain sufficient potential seed pairs, thereby supplementing the scarce alignment signal.

	To begin with, in order to eliminate differences between entity semantics in different languages, we standardize the entity semantics in the cross-lingual knowledge graph to ensure they are presented in the same language form. In this work, they are all presented in English form. Subsequently, these entity semantic information are vectorized through BGE\cite{xiao2023c} to obtain their embedding representation: 
	\begin{equation}
		\begin{aligned}
			{\mathbf{h}_{{e_{1}}}^{s},\mathbf{h}_{{e_{2}}}^{s}...\mathbf{h}_{{e_{n}}}^{s}} = BGE(semantic\{e_1,e_2...e_n\})
		\end{aligned}
	\end{equation}
	where $semantic\{e_1,e_2...e_n\}$ represents the entity's semantic attribute. Next, we consider the structure of the entities in the knowledge graph to obtain the neighbor semantic information of the entities. We aggregate the neighborhood semantic embeddings of entities through the convolution operation of the GCN. This aggregation process utilizes the low-pass filtering characteristics of the convolution operation, which can retain stable correlation features in the neighborhood and filter out outliers and noise. This can reduce the noise interference in neighborhood semantic information to a certain extent and obtain high-quality neighborhood semantic embedding representation:
	\begin{equation}
		\begin{aligned}
			\mathbf{h}_{{e_{i}}}^{n}= \frac{1}{\sqrt{|N_{e_i}||N_{e_j}|}} \sum_{e_j\in N_{e_i}} \mathbf{h}_{{e_{j}}}^{s}
		\end{aligned}
	\end{equation}
	where $N_{e_i}$ represents the neighbor set of $e_i$, and $N_{e_j}$ represents the neighbor set of $e_j$. Then, for entity semantic information and neighborhood semantic information, we construct cosine similarity matrices $\mathbf{M}_{s}^{cos}$ and $\mathbf{M}_{n}^{cos}$. While these matrices reflect the absolute similarity between entities, they have some limitations. Specifically, when the similarity between the source entity and most candidate entities is generally high but lacks clear differentiation, the target entity derived from these matrices may not be the best match. To address this issue, we introduce cross-domain similarity local scaling(CSLS)\cite{lample2018word}. CSLS utilizes the similarity between the entity and its Top-Q nearest neighbors to adjust the similarity between the entity and candidate entities. This adjustment enables the target entity to exhibit a higher weight distribution in comparisons, thereby improving the accuracy of the alignment. we apply CSLS to $\mathbf{M}_{s}^{cos}$ and $\mathbf{M}_{n}^{cos}$, resulting in the adjusted $\mathbf{M}_{s}^{csls}$ and $\mathbf{M}_{n}^{csls}$:
	\begin{equation}
		\begin{aligned}
			\mathbf{M}_{x}^{csls} = CSLS(\mathbf{M}_{x}^{\cos})  \quad x \in \{s, n\}
		\end{aligned}
	\end{equation}

	\begin{equation}
		\begin{aligned}
			 CSLS(\mathbf{M}^{\cos}_{x}) = 2 \mathbf{M}^{\cos}_{x} - \mathbf{M}^{avg_1}_{x} - \mathbf{M}^{avg_2^T}_{x}
		\end{aligned}
	\end{equation}
	
	\begin{equation}
		\begin{aligned}
			\mathbf{M}^{\cos}_{x} = \cos(\mathbf{H}_{x_{1}}, \mathbf{H}_{x_{2}}^{T})
		\end{aligned}
	\end{equation}
	
	\begin{equation}
		\begin{aligned}
			\mathbf{M}_{x_{(i,:)}}^{avg_1} = \frac{1}{Q}  \sum_{j \in \text{Top} Q} \mathbf{M}_{x_{(i, j)}}^{\cos}
		\end{aligned}
	\end{equation}
	
	\begin{equation}
		\begin{aligned}
			\mathbf{M}_{x_{(:, j)}}^{avg_2} = \frac{1}{Q}  \sum_{i \in \text{Top} Q} \mathbf{M}_{x_{(i, j)}}^{\cos}
		\end{aligned}
	\end{equation}
	where $s$ and $n$ represent the semantic attributes of the entity and the semantic attributes of the neighborhood, $\mathbf{M}^{\cos}_{x}$ represents the cosine similarity matrix, $\mathbf{M}_{x_{(i,:)}}^{avg_1}$ represents the average distance between the source entity and its Top-Q nearest neighbors in the candidate entity set and $\mathbf{M}_{x_{(:, j)}}^{avg_2}$ represents the average distance between the candidate entity and its Top-Q nearest neighbors in the source entity set. We combine these two similarity matrices to obtain the neighborhood-level entity semantic similarity matrix: 
	\begin{equation}
		\begin{aligned}
			\mathbf{M}^{sem} = \epsilon \mathbf{M}_{s}^{csls} + (1 - \epsilon) \mathbf{M}_{n}^{csls}
		\end{aligned}
	\end{equation}
	where $\epsilon$ represents hyperparameters for combining two similarity matrices. Next, we select initial potential seed pairs by combining the similarity threshold and bidirectional nearest neighbor strategy: 
	
	\begin{equation}
		\bm{S}_I = \left\{ (e_i, e_j) \left| 
		\begin{array}{l}
			\forall j' \neq j,  \mathbf{M}^{sem}_{ij} > \mathbf{M}^{sem}_{ij'} \\[3pt]
			\forall i' \neq i,  \mathbf{M}^{sem}_{ij} > \mathbf{M}^{sem}_{i'j} \\[3pt]
			\mathbf{M}^{sem}_{ij} > \theta_{sem}  \\[3pt]
			(e_i \notin \bm{S}) \lor (e_j \notin \bm{S})
		\end{array}
		\right. \right\}
	\end{equation}
	where $\bm{S}$ represents the pre-aligned seed set, $\bm{S}_I$ represents the initial potential seed set, $\theta_{sem}$ represents the semantic similarity threshold. Specifically, only when both parties of the entity pair are considered to be each other's nearest neighbors, the neighborhood-level information similarity is greater than $\theta_{sem}$, and neither party of the entity pair is in the pre-aligned seed will they be used as the initial potential seed pair. Finally, we merge the initial potential seed set  with the pre-aligned seed set to obtain the expanded seed set: 
	\begin{equation}
		\begin{aligned}
		\bm{S}_E = \bm{S} \cup \bm{S}_I
		\end{aligned}
	\end{equation}
	where $\bm{S}_E$ represents the expand seed set.  
	
	\subsection{\textbf{Local and Global Awareness Mechanism}}
	
	In this part, we construct a comprehensive local and global awareness mechanism (LGAM) to mine local relation information and global entity information in KG, thereby obtaining a more comprehensive entity embedding representation for better iterative optimization of seed pairs. Fig.\ref{fig3} describes the process details of LGAM, which mainly includes local relation awareness and global entity awareness. Local relation awareness combines the neighborhood information of entities to generate local embeddings of entities. Global entity awareness first builds high-order neighbors based on semantic information and then generates global embeddings of the entity by aggregating the high-order neighbors of the entity. Finally, the embeddings are connected to obtain the final embeddings. Next, we will describe this method in detail. 
	
	\subsubsection{\textbf{Local Relation Awareness}}
	Firstly, by aggregating neighboring entity representations and relation representations, we generate initial local entity embeddings $\mathbf{h}_{e_{i}}^{e}$ and local relation embeddings $\mathbf{h}_{e_{i}}^{r}$: 
	
	\begin{equation}
		\begin{aligned}
			\mathbf{h}_{{e_{i}}}^{e}=\frac{1}{|N_{e_i}|} \sum_{e_j\in N_{e_i}} \mathbf{h}_{{e_j}}
		\end{aligned}
	\end{equation}
	\begin{equation}
		\begin{aligned}
			\mathbf{h}_{{e_{i}}}^{r}=\frac{1}{|R_{e_i}|}\sum_{r_k\in R_{e_i}}\mathbf{h}_{{r_k}}
		\end{aligned}
	\end{equation}
	where $N_{e_i}$ represents the neighbor set of the $e_i$, and $R_{e_i}$ represents the relation set between the $e_i$ and all neighbor entities.We respectively use $\mathbf{h}_{{e_{i}}}^{e}$ and $\mathbf{h}_{{e_{i}}}^{r}$ as the embedding input $\mathbf{h}_{e_{i}}$ of the model to enhance the model's understanding of entities and relationships. Then, the neighborhood information of the entity is weighted and aggregated through local relation awareness attention:
	\begin{equation}
		\begin{aligned}
			\mathbf{h}_{{e_{i}}}^{l+1}=\tanh(\sum_{r_k\in R_{e_i}}\alpha_k(\mathbf{h}_{{e_{j}}}^{l}-2\mathbf{h}_{{r_{k}}}^{T}\mathbf{h}_{{e_{j}}}^{l}\mathbf{h}_{{r_{k}}}))
		\end{aligned}
	\end{equation}
	\begin{equation}
		\begin{aligned}
			\alpha_k = \frac{\exp({\mathbf{v}_1}{\mathbf{h}_{r_k}})}{\sum_{r_{k'}\in R_{e_i}} \exp({\mathbf{v}_1}{\mathbf{h}_{r_{k'}}})}
		\end{aligned}
	\end{equation}
	where $\mathbf{h}_{{e_{i}}}^{l+1}$ represents the local information representation of the $e_i$, $e_j$ represents the neighbor entity corresponding to the $e_i$ under the $r_k$, $\alpha_k$ represents the local relation attention coefficient, and $\mathbf{v}_1$ represents the attention weight parameter. Finally, we stack multiple layers of entity representations to expand the field of view of local embeddings, obtaining a more comprehensive feature expression. 
	\begin{equation}
		\begin{aligned}
			\mathbf{h}_{{e_{i}}}^{local}=[\mathbf{h}_{{e_{i}}}^{0}||\mathbf{h}_{{e_{i}}}^{1}||...||\mathbf{h}_{{e_{i}}}^{l}]
		\end{aligned}
	\end{equation}

	\subsubsection{\textbf{Global Entity Awareness}}
	
	In this part, we consider the global perspective and use the global semantic information on the graph to improve the embedding representation ability of feature diversity entities. We observe that entity semantic attributes offer two key advantages: they are easily accessible and effectively reflect the high-order semantic relationships between entities. Therefore, based on entity semantic information, we select semantic higher-order neighbors and introduce global semantic information into the model to obtain entity embeddings that fully integrate individual entity features and semantic information. To begin, we use the CSLS method to calculate the semantic embedding distance between entities to better capture the semantic similarity differences between entities:
	\begin{equation}
		\begin{aligned}
			\mathbf{M}_{dis}^{csls} = CSLS(cos(\mathbf{H},\mathbf{H}^T))
		\end{aligned}
	\end{equation}
	where $\mathbf{H}\in R^{(N*d)}$ represents the entity semantic embedding matrix. Then, for each entity, we select the top K entities with the highest semantic similarity from the graph to serve as the high-order semantic neighbors of that entity:
	\begin{equation}
		\begin{aligned}
			\bm{E}_{Topk}^{sem} = TopK(\mathbf{M}_{dis}^{csls},K)
		\end{aligned}
	\end{equation}
	where $\bm{E}_{Topk}^{sem}$ represents semantic high-order neighbor set of the entity. By adjusting the value of $K$, we can choose an appropriate number of high-order neighbors, thereby providing more accurate high-frequency features. Similar to the local relation awareness, next, we utilize global entity awareness attention to obtain the global information representation of the entities, where the global entity attention coefficients are as follows:	
	\begin{equation}
		\begin{aligned}
			\beta_{ij} = \frac{\exp({\mathbf{v}_2}{\mathbf{h}_{e_j}})}{\sum_{e_{j^{'}}\in N_{e_i}^{Topk}} \exp({\mathbf{v}_2}{\mathbf{h}_{e_{j^{'}}}})}
		\end{aligned}
	\end{equation}
	where $\mathbf{v}_2$ represents the attention weight parameter, and $N_{e_i}^{Topk}$ represents the high-order neighbor set of the entity obtained through $\mathbf{E}_{Topk}^{sem}$. Next, we perform weighted aggregation on the high-order neighbor information of entities and, at the same time, expand the scope of global information by stacking multiple layers of entity representations. 
	\begin{equation}
		\begin{aligned}
			\mathbf{h}_{e_i}^{l+1}=\tanh(\sum_{e_j\in N_{e_i}}\beta_{ij}\mathbf{h}_{e_j}^{l})
		\end{aligned}
	\end{equation}
	\begin{equation}
		\begin{aligned}
			\mathbf{h}_{{e_{i}}}^{global}=[\mathbf{h}_{{e_{i}}}^{0}||\mathbf{h}_{{e_{i}}}^{1}||...||\mathbf{h}_{{e_{i}}}^{l}]
		\end{aligned}
	\end{equation}
	
	Finally, we obtain the final entity embedding by splicing local entity embeddings and global entity embeddings. 
	
	\begin{equation}
		\begin{aligned}
			\mathbf{h}_{{e_{i}}}^{final}=[\mathbf{h}_{{e_{i}}}^{local}||\mathbf{h}_{{e_{i}}}^{global}]
		\end{aligned}
	\end{equation}

	\begin{algorithm}
		\caption{Threshold nearest neighbor embedding correction strategy}
		\Input{{$\bm{E_1}$} and {$\bm{E_2}$}, {$\mathbf{H}_{fin_{1}}$} and {$\mathbf{H}_{fin_{2}}$}, $\theta_{fin}$, $\bm{S}$}
		\Output{$\bm{S}_{E}$}
		
		{$\mathbf{M}^{fin}$} =  $CSLS$ ($cos$({$\mathbf{H}_{{fin}_{1}}$},{$\mathbf{H}_{fin_{2}}^{T}$}))\;
		${\bm{S}_O} \leftarrow \{\}$\;
		\ForEach{$e_i \in \bm{E_1}$}{
			$e_j \leftarrow \arg\max_{e_{j'} \in \bm{E_2}} \mathbf{M}^{fin}_{ij'}$\;
			\If{$\forall j' \neq j,  \mathbf{M}^{fin}_{ij} > \mathbf{M}^{fin}_{ij'}$ \newline and $\forall i' \neq i,  \mathbf{M}^{fin}_{ij} > \mathbf{M}^{fin}_{i'j}$ \newline and $\mathbf{M}^{fin}_{ij} > \theta_{fin}$ \newline and $(e_i \notin \bm{S}) \lor (e_j \notin \bm{S})$;\newline}{
				$\bm{S}_O \leftarrow \bm{S}_O + \{(e_i, e_j)\}$\;
			}
		}
		$\bm{S}_{E} \leftarrow \bm{S} \cup \bm{S}_O$\;
		$\mathbf{H}_{cor_{1}},\mathbf{H}_{cor_{2}} = Xavier (\mathbf{H}_{fin_{1}},\mathbf{H}_{fin_{2}})$\;
		\Return $\bm{S}_{E}$ 
	\end{algorithm}
	
	\subsection{\textbf{Iterative Optimization}}
	
	As the scale of knowledge graphs expands, the cost of manually labeling seed pairs is increasing. Many models\cite{sun2018bootstrapping,mao2020mraea,zhang2023semi} utilize semi-supervised iterative strategies to expand the training seed set. Although these strategies can select many high-quality potential seed pairs, they all overlook the issue of embedding distortion caused by noisy seed pairs during the training process. We propose the threshold nearest neighbor embedding correction strategy (TNECS) during the semi-supervised iterative optimization phase to address these concerns. 
	
	Specifically, when the iteration round conditions are met, we use $\mathbf{h}^{final}$ obtained from LGAM to construct the embedding similarity matrix.  
	\begin{equation}
		\begin{aligned}
			\mathbf{M}^{fin} = CSLS(cos(\mathbf{H}_{fin_{1}},\mathbf{H}_{fin_{2}}^{T}))
		\end{aligned}
	\end{equation}
	where $\mathbf{H}_{fin}$ represents the embedding matrix for $\mathbf{h}^{final}$. Then, by integrating the similarity threshold and bidirectional nearest neighbor method, we obtain the optimized potential seed pairs and update the expanded seed set: 
	\begin{equation}
		\bm{S}_O = \left\{ (e_i, e_j) \left| 
		\begin{array}{l}
			\forall j' \neq j,  \mathbf{M}^{fin}_{ij} > \mathbf{M}^{fin}_{ij'} \\[3pt]
			\forall i' \neq i,  \mathbf{M}^{fin}_{ij} > \mathbf{M}^{fin}_{i'j}\\[3pt]
			\mathbf{M}^{fin}_{ij} > \theta_{fin}  \\[3pt]
			(e_i \notin \bm{S}) \lor (e_j \notin \bm{S})
		\end{array}
		\right. \right\}
	\end{equation}
	
	\begin{equation}
		\begin{aligned}
			\bm{S}_E = \bm{S} \cup \bm{S}_O
		\end{aligned}
	\end{equation}
	where $\bm{S}_O$ represents the optimized potential seed pairs, $\theta_{fin}$ represents the final emdeddings similarity threshold. Subsequently, we perform an embedding correction operation to reset $\mathbf{H}_{fin}$ to a uniformly distributed initialization state using the Xavier initializer. This strategy aims to mitigate the embedding distortion caused by the noise seeds while alleviating the gradient vanishing problem to a certain extent by maintaining the stability of the activation values and gradients.

	\begin{equation}
		\begin{aligned}
			{\mathbf{H}_{cor_{1}},\mathbf{H}_{cor_{2}}} = Xavier (\mathbf{H}_{fin_{1}},\mathbf{H}_{fin_{2}})
		\end{aligned}
	\end{equation}
	
	Finally, we bring the $\bm{S}_E$ and corrected $\mathbf{H}_{cor}$ into LGAM again for iterative training. The detailed iterative optimization process is presented in Algorithm 1.

	\subsection{\textbf{Model Training}}
	In entity alignment tasks, it is common to use the distance between entities as a measure of whether entities are aligned. The greater the similarity between entities, the smaller their distance, indicating a higher likelihood of being aligned entity pairs. We opt for the Euclidean distance as the standard for measuring the similarity between entities. Specifically, for entities in KG1 and KG2, their distance can be represented as: 
	\begin{equation}
		\begin{aligned}
			d_{(e_i,e_j)}=||\mathbf{h}_{{e_{i}}}^{final}-\mathbf{h}_{{e_{j}}}^{final}||_{L2}
		\end{aligned}
	\end{equation}
	where $\mathbf{h}_{{e_{i}}}^{final}$ and $\mathbf{h}_{{e_{j}}}^{final}$ represent the entity embeddings obtained through LGAM. Next, we use the loss function based on $LogSumExp$\cite{mao2021boosting} to calculate the loss of entity alignment and minimize the loss function through the optimization algorithm.  
	\begin{align}
		L &= \sum_{(e_i,e_j)\in \bm{S}_{E}}\log\Bigg[1\!+\!\sum_{e_{j'}\in E_2} \exp\big(\gamma(\lambda \!+\! d_{(e_i,e_j)} \!-\! d_{(e_i,e_{j'})})\big)\Bigg] \notag \\
		&+ \sum_{(e_i,e_j)\in \bm{S}_{E}}\log\Bigg[1\!+\!\sum_{e_{i'}\in E_1} \exp\big(\gamma(\lambda \!+\! d_{(e_i,e_j)} \!-\! d_{(e_{i'}\!,e_j)})\big)\Bigg]
	\end{align}

	Where $\bm{S}_{E}$ represents the expanded seed set used as the positive sample set, $\gamma$ is the scaling factor, and $\lambda$ is the margin hyperparameter.
	
	\subsection{\textbf{Complexity Analysis}}
	To gain a comprehensive understanding of the model's performance, we conducted a time complexity analysis focusing on three key components: seed expansion, local and global awareness, and iterative optimization. In both the seed expansion and iterative optimization stages, constructing the similarity matrix is pivotal for determining the time complexity. The complexity of this operation is $\mathcal{O}(|E_1| \times |E_2|)$, where $E_1$ and $E_2$ represent the number of entities in the source and target knowledge graphs, respectively. The local and global awareness stages are primarily governed by the calculation of attention coefficients and neighbor information aggregation, which depend on the number of first-order neighbors and high-order semantic neighbors of the entities. The time complexity for this stage is $\mathcal{O}(|E| \times (|R| + |K|))$, where $E$ denotes the total number of entities, $R$ represents the number of relations, and $K$ signifies the number of high-order semantic neighbors. In summary, the time complexity of SE-GNN is $\mathcal{O}(|E_1| \times |E_2| + |E| \times (|R| + |K|))$.
	
	\section{EXPERIMENT}
	\label{section: section4}
	\subsection{\textbf{Experiment Setup}}
	\subsubsection{\textbf{Datasets}}
	We conducted experiments of SE-GNN on three widely used public datasets. The statistical data of these datasets are shown in Table \ref{table1}. 
	
	\textbf{DBP15K}\cite{sun2017cross}: Comprising three multilingual datasets DBP15K$_{\text{ZH-EN}}$, DBP15K$_{\text{JA-EN}}$ and DBP15K$_{\text{FR-EN}}$. Each dataset contains around 20,000 entity pairs, with 15,000 pairs already labeled for alignment and available for training or testing. There are about 5,000 additional entity pairs where many are aligned but not labeled. 
	
	\textbf{SRPRS}\cite{guo2019learning}: SRPRS$_{\text{EN-FR}}$ and SRPRS$_{\text{EN-DE}}$ are two cross-lingual datasets in SRPRS. Both consist of 15,000 entity pairs, all labeled for alignment. Compared to DBP15K, SRPRS is sparser, and the degree distribution of nodes aligns more with real-world knowledge graphs.
	
	\textbf{DWY100K}\cite{sun2018bootstrapping}: It consists of two single-language datasets, DWY100K$_{\text{WD-DBP}}$ and DWY100K$_{\text{YG-DBP}}$. Each dataset contains 100,000 aligned entity pairs. Compared with small-scale datasets such as DBP15K, the number of entities in DWY100K has increased significantly, making it more suitable for verifying the efficiency and robustness of the model in large-scale scenarios.
	
	\begin{table}[!htbp]
		\centering
		\renewcommand\arraystretch{1.5}
		\caption{Analysis of the DBP15K, SRPRS and DWY100K dataset.}
		\begin{tabular}{c@{\setlength{\tabcolsep}{0pt}}c|ccccc}
			\specialrule{1.2pt}{0pt}{0pt} 
			\multicolumn{2}{c|}{\multirow{1}{*}{Datasets}}   & Entities   & Relations   & Triples \\
			\specialrule{1.2pt}{0pt}{0pt} 
			\multirow{2}{*}{DBP15K$_{\mathrm{ZH-EN}}$}  & Chinese  & 19388  & 1701 & 70414     \\
			~           & English   & 19572  & 1323 & 9514     \\
			\multirow{2}{*}{DBP15K$_{\mathrm{JA-EN}}$}  & Japanese    & 19818  & 1299 & 77214     \\
			~           & English   & 19780  & 1153 & 93484     \\
			\multirow{2}{*}{DBP15K$_{\mathrm{FR-EN}}$}  & French   & 19661  & 903 & 105998     \\
			~           & English   & 19993  & 1208 & 115722     \\
			\hline
			\multirow{2}{*}{SRPRS$_{\mathrm{EN-FR}}$}  & English    & 15000  & 221 & 36508     \\
			~           & French   & 15000  & 177 & 33532     \\
			\multirow{2}{*}{SRPRS$_{\mathrm{EN-DE}}$}  & English   & 15000  & 222 & 38363     \\
			~           & German   & 15000  & 120 & 37377     \\ 
			\hline
			\multirow{2}{*}{DWY100K$_{\mathrm{WD-DBP}}$}  & Wikipedia    & 100000  & 220 & 448774     \\
			~           & DBpdia   & 100000  & 330 & 463294     \\
			\multirow{2}{*}{DWY100K$_{\mathrm{YG-DBP}}$}  & YAGO3   & 100000  & 31 & 502563     \\
			~           & DBpdia   & 100000  & 302 & 428952     \\ 
			\specialrule{1.2pt}{0pt}{0pt} 
			\label{table1}
		\end{tabular}
	\end{table}

	\subsubsection{\textbf{Parameter}}
	
	We employ a greedy search strategy to select the best hyperparameters. For each individual hyperparameter, we sequentially evaluate its possible values to identify the optimal option. This process continues until the best value is determined for all hyperparameters. Specifically, we consider the following ranges of hyperparameter values. The dimension of entity embedding in \{50, 100, 150, 200\}, the dimension of relation embedding in \{50, 100, 150, 200\}, the number of nearest neighbors of CSLS in \{5, 10, 15, 20\}, the number of semantic higher-order neighbors in \{5, 15, 25, 35\}, the layers of GNN number in \{1, 2, 3, 4\}, $\epsilon$ of the matrix fusion parameters in \{0.3, 0.4, 0.5, 0.6\}, $\theta_{sem}$ of the semantic similarity threshold in \{0.01, 0.02, 0.03, 0.04\}, $\theta_{fin}$ of the final emdeddings similarity threshold in \{0.03, 0.05, 0.07, 0.09\}, the learning rate in \{0.001, 0.005, 0.01\}, the optimization round interval \{10, 20, 30, 40\}. The parameters used in the final experiments were as follows:
	
	The embedding dimensions of entity and relation are both 100, the number of nearest neighbors Q for CSLS is 15 , the number of semantic high-order neighbors K is 15, the depth l of GNN is 2, the matrix fusion parameters $\epsilon$ is 0.5, the semantic similarity threshold $\theta_{sem}$ is 0.01, and the final embeddings similarity threshold $\theta_{fin}$ is 0.05. We use RMSprop to optimize the model with a learning rate of 0.01. The optimization round interval of TNECS is set to 30, the Xavier initializer is used for entity embedding correction, and TNECS is updated 3 times. 
	
	In our experiments, we partitioned the datasets as follows: 30\% of the seed pairs were designated as the training set, 10\% as the validation set and 60\% as the test set. We employed ten-fold cross-validation to ensure robust evaluation and implemented an early stopping strategy to avoid over-optimization of the model on the training set. The final experimental results are the average of ten training runs conducted on an NVIDIA 3090 with 24GB of memory.

	\subsection{\textbf{Baselines}}

	We have divided SE-GNN into two versions, SE-GNN (tradi) and SE-GNN (semi), in order to more comprehensively compare and analyze the components of SE-GNN. 
	
	SE-GNN (tradi) does not involve semi-supervised iterative training but only utilizes entity embeddings obtained from LGAM for alignment. SE-GNN (semi) is an iterative strategy. It adds seed expansion and seed optimization based on SE-GNN (tradi). We compare these two versions with ten baseline models, which can also be categorized into traditional entity alignment methods and semi-supervised entity alignment methods.

	\subsubsection{\textbf{Traditional Entity Alignment Methods}}
	\begin{itemize}
		\item GCN-Align\cite{wang2018cross} aggregates neighborhood information of entities through graph convolution operations.
	\end{itemize}
	
	\begin{itemize}
		\item MuGNN\cite{cao2019multi} uses an attention mechanism to simultaneously aggregate relation information and attribute information.
	\end{itemize}
	
	\begin{itemize}
		\item AliNet\cite{sun2020knowledge} aggregates multi-hop neighbor information through an attention mechanism and uses a gating mechanism to combine the representations of multiple aggregation functions.
	\end{itemize}
	
	\begin{itemize}
		\item Dual-AMN\cite{mao2021boosting} uses proxy matching vectors to change the calculation between nodes into the calculation between nodes and proxy matching vectors, reducing the computational complexity.
	\end{itemize}

	\subsubsection{\textbf{Semi-Supervised Entity Alignment Methods}}
	
	\begin{itemize}
		\item BootEA\cite{sun2018bootstrapping} marks potentially aligned entities as training data in an iterative manner and uses an alignment editing strategy to reduce error accumulation.
	\end{itemize}
	
	\begin{itemize}
		\item GALA\cite{zhang2023semi} uses entity embedding to build global features. Align entities in the graph by forcing global features to match each other and incrementally update entity embeddings by aggregating local information from other networks.
	\end{itemize}
	
	\begin{itemize}
		\item MRAEA\cite{mao2020mraea} combines the incoming and outgoing neighbors of entities and the meta-semantics of connection relationships to represent entities while filtering aligned entities through a bidirectional iterative strategy.
	\end{itemize}
	
	\begin{itemize}
		\item TransEdge\cite{sun2019transedge} combines the embedded representations of entities and relationships to update the relationships between relationship entities to obtain edge diversity.
	\end{itemize}
	
	\begin{itemize}
		\item RANM\cite{cai2022semi} is based on relational matching to find the larger range and higher confidence neighborhoods of aligned entities. 
	\end{itemize}
	
	\begin{itemize}
		\item DATTI\cite{mao2022effective} uses the adjacency and internal correlation isomorphism of KG to propose an EA decoding algorithm based on third-order tensor isomorphism to enhance the EA decoding process.
	\end{itemize}
	
	The results of the baseline models are mostly from their respective papers or code reproductions, and their hyperparameters are consistent with the original descriptions. In addition, some of the results come from the implementation of Dual-AMN\cite{mao2021boosting}.
	
\begin{table*}[]
	\centering
	\renewcommand\arraystretch{1.1}
	\setlength{\tabcolsep}{3.8pt}
	\caption{Performance comparison of SE-GNN with baseline models on DBP15K.}
	\begin{tabular}{llccccccccccc}
		\specialrule{1.2pt}{0pt}{0pt}
		& & \multicolumn{3}{c}{DBP15K$_{\mathrm{ZH-EN}}$}  & & \multicolumn{3}{c}{DBP15K$_{\mathrm{JA-EN}}$} & & \multicolumn{3}{c}{DBP15K$_{\mathrm{FR-EN}}$}\\ 
		\cline{3-5} \cline{7-9} \cline{11-13}
		& & Hit@1  & Hit@10  & MRR  &  & Hit@1  & Hit@10  & MRR  &  & Hit@1  & Hit@10  & MRR  \\
		\specialrule{1.2pt}{0pt}{0pt}
		\multirow{5}{*}{\centering tradi} 
		& GCN-Align    & 42.33 & 74.62 & 0.557 && 41.34 & 75.63 & 0.549 && 40.76 & 76.14 & 0.527 \\
		& MuGNN       & 49.40 & 84.40 & 0.611 && 50.10 & 85.70 & 0.621 && 49.60 & 87.00 & 0.621 \\
		& Alinet      & 53.90 & 82.60 & 0.628 && 54.90 & 83.10 & 0.645 && 55.20 & 85.20 & 0.657 \\
		& Dual-AMN    & 73.10 & 92.30 & 0.799 && 72.60 & 92.70 & 0.799 && 75.60 & 94.80 & 0.827 \\
		\cline{2-13}
		& SE-GNN(tradi) & \textbf{74.44} & \textbf{93.69} & \textbf{0.815} && \textbf{76.10} & \textbf{95.14} & \textbf{0.830} && \textbf{79.51} & \textbf{95.92} & \textbf{0.856} \\
		\hline
		\multirow{6}{*}{\centering semi} 
		& BootEA      & 62.94 & 84.75 & 0.703 && 62.23 & 85.39 & 0.701 && 65.30 & 87.44 & 0.731 \\
		& GALA        & 56.33 & 81.11 & 0.646 && 56.83 & 81.78 & 0.652 && 58.09 & 84.06 & 0.669 \\ 
		& MRAEA       & 75.70 & 92.98 & 0.827 && 75.78 & 93.38 & 0.826 && 78.09 & 94.81 & 0.849 \\
		& TransEdge   & 73.50 & 91.90 & 0.801 && 71.90 & 93.20 & 0.795 && 71.00 & 94.10 & 0.796 \\ 
		& RANM        & 79.01 & 89.08 & 0.825 && 91.59 & 95.30 & 0.929 && 92.43 & 96.24 & 0.937 \\
		& DATTI       & 83.50 & 95.30 & 0.880 && 83.60 & 96.90 & 0.884 && 87.30 & 97.90 & 0.913 \\
		\cline{2-13}
		& SE-GNN(semi) & \textbf{96.34} & \textbf{98.95} & \textbf{0.973} && \textbf{97.31} & \textbf{99.34} & \textbf{0.981} && \textbf{98.04} & \textbf{99.60} & \textbf{0.986} \\
		\specialrule{1.2pt}{0pt}{0pt}
		\multicolumn{13}{@{}l}{\footnotesize \textit{Results are taken from respective papers or code reproductions.}} \\ 
		\label{table2}
	\end{tabular}
\end{table*}
\begin{table*}[]
	\centering
	\renewcommand\arraystretch{1.2}
	\setlength{\tabcolsep}{3.6pt}
	\caption{Performance comparison of SE-GNN with baseline models on SRPRS and DWY100K.}
	\begin{tabular}{llcccccccccccccccc}
		\specialrule{1.3pt}{0pt}{0pt}
		& & \multicolumn{3}{c}{SRPRS$_{\mathrm{EN-FR}}$}  && \multicolumn{3}{c}{SRPRS$_{\mathrm{EN-DE}}$} && \multicolumn{3}{c}{DWY100K$_{\mathrm{WD-DBP}}$} && \multicolumn{3}{c}{DWY100K$_{\mathrm{YG-DBP}}$} \\ 
		\cline{3-5} \cline{7-9} \cline{11-13} \cline{15-17}
		& & Hit@1  & Hit@10  & MRR  && Hit@1  & Hit@10  & MRR  && Hit@1  & Hit@10  & MRR  && Hit@1  & Hit@10  & MRR  \\
		\specialrule{1.2pt}{0pt}{0pt}
		\multirow{5}{*}{\centering tradi} 
		& GCN-Align & 24.53 & 52.46 & 0.341 && 38.73 & 60.31 & 0.469 && 50.63 & 77.37 & 0.613 && 59.74 & 83.27 & 0.681 \\
		& *MuGNN     & 13.10 & 34.20 & 0.208 && 24.50 & 43.10 & 0.310 && 60.40 & 89.40 & 0.701 && 73.90 & 93.70 & 0.810 \\
		& *Dual-AMN  & 45.20 & 74.80 & 0.552 && 59.10 & 82.00 & 0.670 && 79.60 & 95.20 & 0.848 && 86.60 & 97.70 & 0.907 \\
		\cline{2-17}
		& SE-GNN(tradi) & \textbf{69.48} & \textbf{89.63} & \textbf{0.766} && \textbf{77.85} & \textbf{92.20} & \textbf{0.831} && \textbf{91.67} & \textbf{98.18} & \textbf{0.942} && \textbf{98.01} & \textbf{99.81} & \textbf{0.988} \\
		\hline
		\multirow{6}{*}{\centering semi} 
		& *BootEA    & 36.50 & 64.90 & 0.460 && 50.30 & 73.20 & 0.580 && 74.80 & 89.80 & 0.801 && 76.10 & 89.40 & 0.808 \\
		& *MRAEA     & 46.00 & 76.80 & 0.559 && 59.40 & 81.80 & 0.666 && 79.40 & 93.00 & 0.856 && 81.90 & 95.10 & 0.875 \\
		& *TransEdge & 40.00 & 67.50 & 0.490 && 55.60 & 75.30 & 0.630 && 78.80 & 93.80 & 0.824 && 79.20 & 93.60 & 0.832 \\ 
		\cline{2-17}
		& SE-GNN(semi) & \textbf{94.41} & \textbf{96.89} & \textbf{0.952} && \textbf{94.83} & \textbf{97.22} & \textbf{0.956} && \textbf{99.28} & \textbf{99.79} & \textbf{0.995} && \textbf{99.94} & \textbf{99.98} & \textbf{0.999} \\
		\specialrule{1.2pt}{0pt}{0pt}
		\multicolumn{17}{@{}l}{\footnotesize \textit{“*” marks the results obtained from Dual-AMN\cite{mao2021boosting}; other results are taken from respective code reproductions.}} \\ 
		\label{table3}
	\end{tabular}
\end{table*}

	\subsection{\textbf{Evaluation Metrics}}
	In entity alignment, choosing appropriate evaluation indicators can objectively evaluate the effectiveness of the model in solving entity alignment problems. We use $Hits@K$ and MRR as the evaluation indicators of the model. 
	
	\textbf{Hits@k} is a measure of whether there are correctly aligned entities among the top k predicted entities in the ranking results. The formula is as follows:
	\begin{equation}
		\begin{aligned}
			Hits@K = \frac{1}{|\bm{S_t}|}\sum_{e\in \bm{S_t}}|rank_e\le K|
		\end{aligned}
	\end{equation}
	where $\bm{S_t}$ represents the test seed set, $rank_e\le K$ represents the ranking position of the correctly aligned entity after sorting. If the ranking is less than K, the result is counted as 1. The larger the value of $Hits@K$, the better the effect of the model.
	
	\textbf{MRR} measures the reciprocal of the highest ranking of each reference entity in the alignment results and then averages the reciprocal rankings of all predicted entities, as follows:
	\begin{equation}
		\begin{aligned}
			MRR = \frac{1}{|\bm{S_t}|}\sum_{e\in \bm{S_t}}\frac{1}{rank_e}
		\end{aligned}
	\end{equation}
	The value range of MRR is between [0, 1]. The closer the value is to 1, the better the effect of the model.

	\subsection{\textbf{Experiments Result and Analyses}}
	 Table \ref{table2} and Table \ref{table3} shows the results of SE-GNN comparing the baseline model on DBP15K, SRPRS and DWY100K. Across all metrics and datasets, SE-GNN has the best performance regardless of traditional and semi-supervised methods.
	
	In the DBP15K dataset, SE-GNN's Hits@1 exceeds 96\%, Hits@10 exceeds 98\%, and MRR exceeds 0.97, showing significant effect advantages. In the SRPRS dataset, SE-GNN's performance improvement is particularly prominent. Compared with the baseline model, SE-GNN's improvement in Hits@1 index exceeds 35\%. This significant improvement is mainly due to SE-GNN's fusion of high-order semantic information. In the SRPRS dataset, the relationship between entities is relatively sparse, and it is difficult to fully describe the entity only by relying on local information. However, the SE-GNN model significantly expands the information reception range of the entity by incorporating the entity's higher-order semantic information. This in-depth mining and utilization of global semantic information enables SE-GNN to capture richer connections between entities, thus achieving significant performance improvements. In the large-scale dataset DWY100K, SE-GNN's Hits@1, Hits@10 and MRR indicators all exceed 99\%, showing strong generalization ability. In particular, on the sub-dataset DWY100K$_{\mathrm{YG-DBP}}$, SE-GNN achieved near-perfect results, further proving its applicability on large-scale datasets.

	\begin{table*}[]
		\centering
		\renewcommand\arraystretch{1.2}
		\setlength{\tabcolsep}{3pt}
		\caption{Ablation experiment results of different components of SE-GNN (tradi).}
		\begin{tabular}{lccccccccccccccccccccc}
			\specialrule{1.2pt}{0pt}{0pt}
			& \multicolumn{2}{c}{DBP$_{\mathrm{ZH-EN}}$} 
			&& \multicolumn{2}{c}{DBP$_{\mathrm{JA-EN}}$} 
			&& \multicolumn{2}{c}{DBP$_{\mathrm{FR-EN}}$} 
			&& \multicolumn{2}{c}{SRPRS$_{\mathrm{EN-FR}}$} 
			&& \multicolumn{2}{c}{SRPRS$_{\mathrm{EN-DE}}$} 
			&& \multicolumn{2}{c}{DWY$_{\mathrm{WD-DBP}}$} 
			&& \multicolumn{2}{c}{DWY$_{\mathrm{YG-DBP}}$} \\ 
			\cline{2-3} \cline{5-6} \cline{8-9} \cline{11-12} \cline{14-15} \cline{17-18} \cline{20-21}
			& Hit@1 & MRR && Hit@1 & MRR && Hit@1 & MRR && Hit@1 & MRR && Hit@1 & MRR && Hit@1 & MRR && Hit@1 & MRR \\
			\specialrule{1.2pt}{0pt}{0pt}
			GCN-Align    & 42.33 & 0.571 && 41.34 & 0.549 && 40.76 & 0.527 && 24.53 & 0.341  && 38.73 & 0.469 && 50.63 & 0.613 && 59.74 & 0.681 \\
			+LRA         & 70.51 & 0.780 && 69.62 & 0.777 && 73.04 & 0.808 && 43.47 & 0.533 && 57.35 & 0.653 && 81.64 & 0.869 && 87.04 & 0.911 \\
			+GEA         & \textbf{74.44} & \textbf{0.815} && \textbf{76.10} & \textbf{0.830} && \textbf{79.51} & \textbf{0.856} && \textbf{69.48} & \textbf{0.766} && \textbf{77.85} & \textbf{0.831} && \textbf{91.67} & \textbf{0.942} && \textbf{98.01} & \textbf{0.988} \\
			\specialrule{1.2pt}{0pt}{0pt}
			\label{table4}
		\end{tabular}
	\end{table*}
	\begin{table*}[]
		\centering
		\renewcommand\arraystretch{1.2}
		\setlength{\tabcolsep}{3pt}
		\caption{Ablation experiment results of different components of SE-GNN (semi).}
		\begin{tabular}{lccccccccccccccccccccc}
			\specialrule{1.2pt}{0pt}{0pt}
			& \multicolumn{2}{c}{DBP$_{\mathrm{ZH-EN}}$} 
			&& \multicolumn{2}{c}{DBP$_{\mathrm{JA-EN}}$} 
			&& \multicolumn{2}{c}{DBP$_{\mathrm{FR-EN}}$} 
			&& \multicolumn{2}{c}{SRPRS$_{\mathrm{EN-FR}}$} 
			&& \multicolumn{2}{c}{SRPRS$_{\mathrm{EN-DE}}$} 
			&& \multicolumn{2}{c}{DWY$_{\mathrm{WD-DBP}}$}
			&& \multicolumn{2}{c}{DWY$_{\mathrm{YG-DBP}}$} \\ 
			\cline{2-3} \cline{5-6} \cline{8-9} \cline{11-12} \cline{14-15} \cline{17-18} \cline{20-21}
			& Hit@1 & MRR && Hit@1 & MRR && Hit@1 & MRR && Hit@1 & MRR && Hit@1 & MRR && Hit@1 & MRR && Hit@1 & MRR \\
			\specialrule{1.2pt}{0pt}{0pt}
			SE-GNN(tradi)    & 74.44 & 0.815 && 75.77 & 0.827 && 78.98 & 0.852 && 69.48 & 0.766 && 77.85 & 0.831 && 91.67 & 0.942 && 98.01 & 0.988 \\
			+BIS             & 82.27 & 0.871 && 83.37 & 0.882 && 86.47 & 0.906 && 74.99 & 0.808 && 82.02 & 0.860 && 96.07 & 0.973 && 99.02 & 0.994 \\
			+TNECS           & 84.51 & 0.887 && 85.63 & 0.899 && 88.75 & 0.921 && 78.39 & 0.834 && 84.86 & 0.883 && 97.03 & 0.979 && 99.45 & 0.997 \\
			+NSI             & \textbf{96.34} & \textbf{0.973} && \textbf{97.31} & \textbf{0.981} && \textbf{98.04} & \textbf{0.986} && \textbf{94.41} & \textbf{0.952} && \textbf{94.83} & \textbf{0.956} && \textbf{99.28} & \textbf{0.995} && \textbf{99.94} & \textbf{0.999} \\
			\specialrule{1.2pt}{0pt}{0pt}
			\label{table5}
		\end{tabular}
	\end{table*}

	We believe that the model's excellent performance is mainly due to the following three key factors: 
	
	\subsubsection{\textbf{Seed expansion based on neighborhood-level semantic information}}
	SE-GNN combines semantic attributes and structural features to expand the seed set and obtain more alignment signals, which can capture richer latent semantic information in the graph while reducing the risk of overfitting.

	\subsubsection{\textbf{Joint propagation of local and global information}}
	SE-GNN aggregates local relational information while also constructing semantic high-order neighbors to propagate global information, thereby enhancing sensitivity to diverse information and effectively alleviate the structural heterogeneity problem of KG.
	
	\subsubsection{\textbf{Embedded correction in iterative optimization}}
	SE-GNN adopts an embedding correction strategy in semi-supervised iterative training to eliminate the embedding distortion caused by noise seeds. It ensures the stability and accuracy of the embeddings during training, thereby improving the model's overall performance.

	\subsection{\textbf{Ablation Studies}}
	Through the above experiments, we proved SE-GNN's overall effectiveness. To evaluate the effectiveness of each component in SE-GNN, we conducted ablation experiments on DBP15K, SRPRS and DWY100K from SE-GNN (tradi) and SE-GNN (semi), respectively.
	
	\subsubsection{\textbf{Ablation Experiments in Traditional Methods}}
	SE-GNN (tradi) obtains local and global information through local relation awareness and global entity awareness to improve the alignment effect. In order to verify the effectiveness of each component, we use GCN-Align\cite{wang2018cross} as the initial encoder and gradually add these two components. The experimental results are shown in Table \ref{table4}.
	
	Compared to the GCN-Align, the local relation awareness module (LRA) uniquely models the neighbors between entities and the relations between neighbors. The global entity awareness module (GEA) introduces global information to alleviate the structural heterogeneity of KG and obtain a more accurate entity representation.
	
	According to the experimental data in Table \ref{table4}, adding these two components each significantly enhanced the model's performance, indicating the effectiveness of our model in incorporating local information while improving entity distinctiveness through global information. Introducing the global entity awareness module, particularly on the SRPRS dataset, led to an improvement of over 22\% in the model's performance. We attribute this success to the sparse distribution of entity relationships in the SRPRS dataset, where relying solely on local entity information is insufficient for adequate description. However, by introducing semantic higher-order neighbors, the global entity awareness module enables entities to access more information, resulting in a significant performance boost in the model. This further validates the importance of the global entity awareness module and its effectiveness in handling sparse entity relationships.

	\begin{figure*}[!t]
		\centering
		\subfloat[]{\includegraphics[width=0.8\columnwidth]{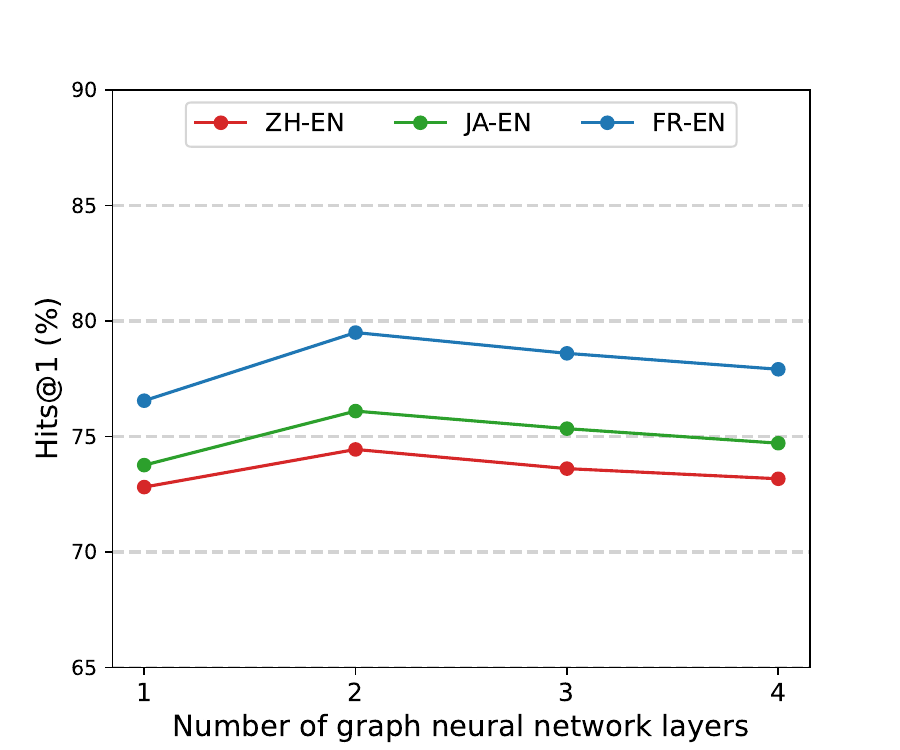}%
			\label{tradi_L}}
		\hfil
		\subfloat[]{\includegraphics[width=0.8\columnwidth]{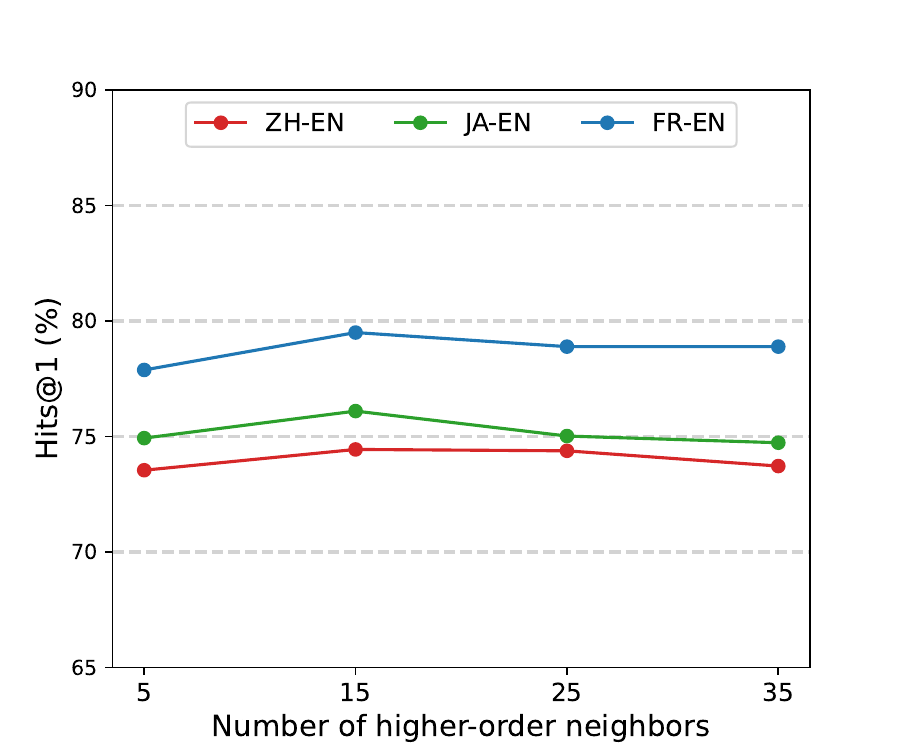}%
			\label{tradi_K}}
		\caption{The effect of different number of neural network layers (a) and number of high-order neighbors (b) on SE-GNN (tradi). }
		\label{tradi}
	\end{figure*}
	\begin{figure*}[!t]
		\centering
		\subfloat[]{\includegraphics[width=0.8\columnwidth]{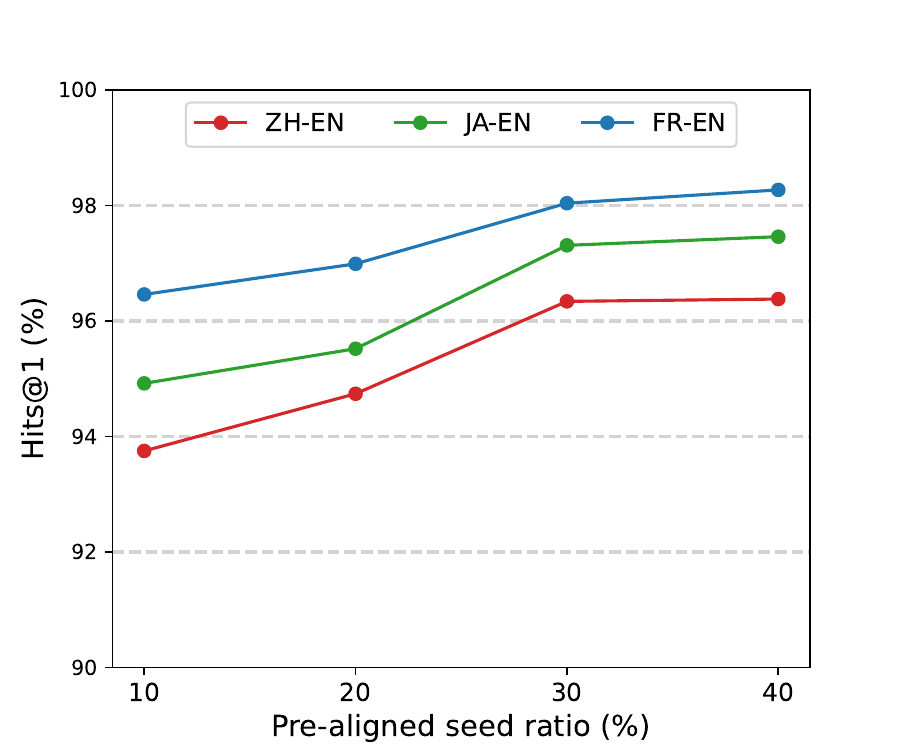}%
			\label{semi_seed}}
		\hfil
		\subfloat[]{\includegraphics[width=0.8\columnwidth]{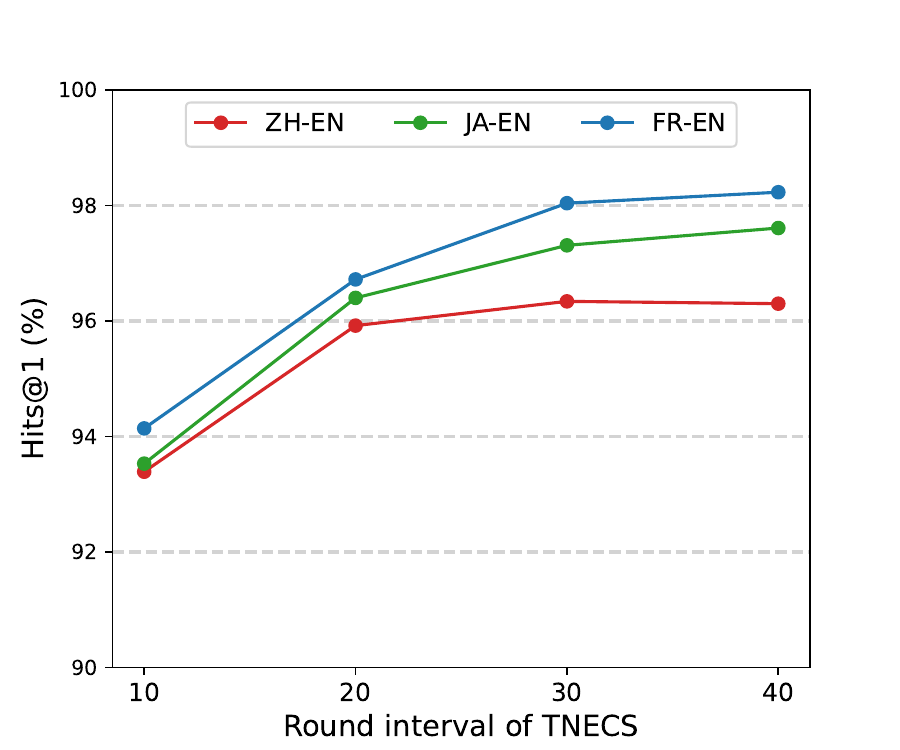}%
			\label{semi_epoch}}
		\caption{The effect of different seed ratios (a) and optimization round intervals (b) on SE-GNN (semi). }
		\label{semi}
	\end{figure*}

	\subsubsection{\textbf{Ablation Experiments in Semi-Supervised Methods}}
	SE-GNN (semi) utilizes neighborhood-level semantic information to build initial potential seed pairs for seed expansion and eliminate the embedding distortion through the threshold nearest neighbor embedding correction strategy. We constructed the following ablation experiment to explore the effects of the two improvements.
	
	BIS means that based on SE-GNN (tradi), bidirectional iterative strategy\cite{mao2020mraea} is used for semi-supervised training. TNECS means replacing the bidirectional iterative strategy and using our threshold nearest neighbor embedding correction strategy for semi-supervised training. NSI means that based on TNECS, the initial potential seed pairs constructed by neighborhood-level semantic information are finally added to enhance alignment.
	
	Table \ref{table5} shows that compared with the bidirectional iterative strategy, the performance of TNECS has been significantly improved. It shows that the entity embedding correction method can eliminate the embedding distortion caused by the selected noise seed pairs during the training process, thereby guiding the model in the right direction. In addition, the addition of initial potential seed pairs also exposes the model to more semantic association information and introduces more alignment signals, allowing for better alignment.

	\subsection{\textbf{Robustness Analysis}}
	In this section, we analyze the selection of hyperparameters in detail to verify the model's robustness. To more accurately analyze the role of hyperparameters in different modules, we also conducted related hyperparameter experiments based on the two dimensions of SE-GNN (tradi) and SE-GNN (semi).
	
	\subsubsection{\textbf{The Impact of Hyper-Parameters in Traditional Modules}}
	In order to study the impact of relevant hyperparameters of the LGAM module, we conducted experiments on the neural network layers and the number of semantic high-order neighbors of SE-GNN (tradi) on the DBP15K data set. Under the premise that all hyperparameters are set to their optimal values, we separately set the number of neural network layers l to {1, 2, 3, 4} and the number of semantic high-order neighbors K to {5, 15, 25, 35}. Fig.\ref{tradi_L} and Fig.\ref{tradi_K} show that when the value of l is 2 and K is 15, the model effect reaches the best, respectively. However, it is worth noting that even under different parameter values, the change in model effect is not large, and the model's performance always remains high. This phenomenon shows that SE-GNN has low dependence on parameters, can achieve good performance under different parameter configurations, and is robust.	
	
	\begin{figure}
		\centering
		\includegraphics[width=1.05\linewidth]{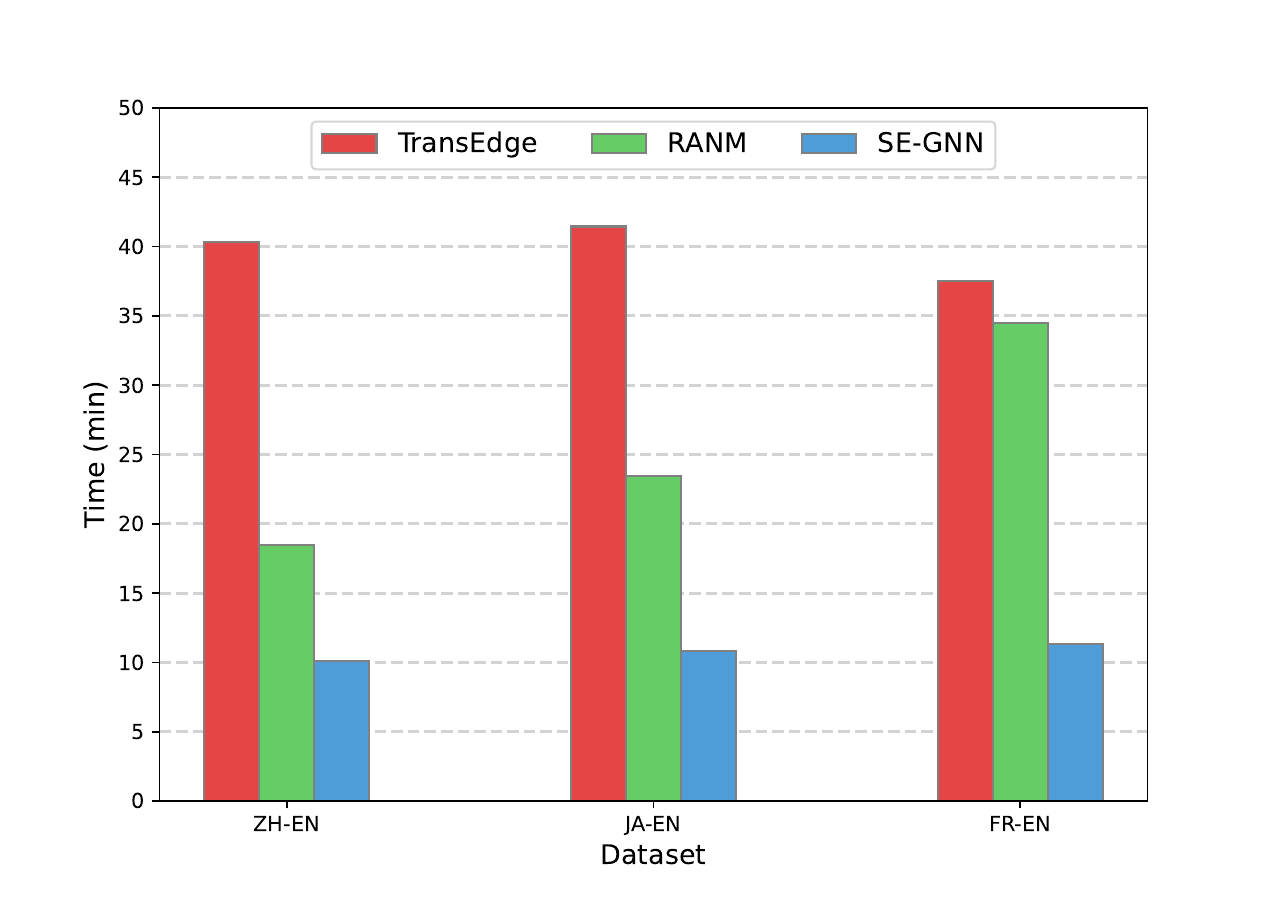}
		\caption{The time costs of TransEdge, RANM, and SE-GNN on the DBP15K dataset. }
		\label{time}
	\end{figure}

	\subsubsection{\textbf{The Impact of Hyper-Parameters in Semi-Supervised Modules}}
	Although most entity alignment methods usually use 30\% of the entity set as the seed set, for larger knowledge graphs, pre-labeling these data also requires higher costs. At the same time, during the iterative training process, the embeddings obtained in different training rounds are different, and the effects of these embeddings on TNECS are also different. In order to study the impact of pre-aligned seed set proportion and optimization round interval on the overall effect, we conducted two tests on the DBP15K data set, namely experiments on different proportions of pre-aligned seed sets and experiments on different optimization round intervals, to study their impact on the SE-GNN (semi) effect.
	
	Under the premise of ensuring that all other hyperparameters are set to their optimal values, we varied the proportion of the pre-aligned seed set from 10\% to 40\% in increments of 10\%. Analyzing the results in Fig.\ref{semi_seed}, we observed that different proportions of seed sets did not significantly change the model effect, and all achieved excellent results. This fully demonstrates that SE-GNN has a low dependence on pre-aligned seed sets. We believe that the reason for achieving such an effect is that SE-GNN obtains sufficient alignment information by selecting potential seed pairs through neighborhood-level entity semantic information, thus reducing the reliance on pre-aligned seed sets. 
	
	When verifying the impact of the interval of optimization rounds, we set TNECS to be performed every 10 to every 40 rounds, with a step size of 10. It can be seen from the experimental results in Fig.\ref{semi_epoch} that when the optimization round interval increases from 10 to 30, the effect of the model gradually increases. However, when it increases from 30 to 40, the slope of the effect increases significantly slows down and is almost the same. This is because potential seed pairs are generated from trained entity embeddings. If the optimization round interval is set too low, the model is not fully trained; if the optimization round interval is too high, entity embedding is prone to overfitting. Both situations will affect the effectiveness of TNECS.

	\subsection{\textbf{Performance Analysis}}
	
	Fig.\ref{time} shows the total running time of SE-GNN and existing EA methods when achieving the best results on the DBP15K dataset. Overall, SE-GNN shows a certain time advantage. Specifically, during the preprocessing stage, SE-GNN takes about 31 seconds to complete seed expansion and 13 seconds to complete high-level neighbor selection. After entering the training phase, SE-GNN takes about 2 to 4 seconds to complete a round of training. This depends on the number of potential seed pairs. Due to the introduction of potential seed pairs, SE-GNN can enter the iterative optimization phase after completing about 30 training rounds. In the iterative optimization phase, each execution of the threshold nearest neighbor embedding correction strategy (TNECS) takes about 18 seconds. Generally, SE-GNN requires about three iterative optimizations to achieve the best results, with the average total training time being approximately 9 to 11 minutes. The seed expansion and iterative optimization stages take a significant amount of time, as both stages require calculating the similarity matrix between entities, which involves substantial computations. However, SE-GNN obtains a large number of potential seed pairs through seed expansion, allowing each training round to utilize more information. This richness of information significantly improves the convergence speed of the model, enabling rapid convergence in fewer training rounds and making the overall training process more efficient.
	
	\begin{figure}
		\centering
		\includegraphics[width=1.05\linewidth]{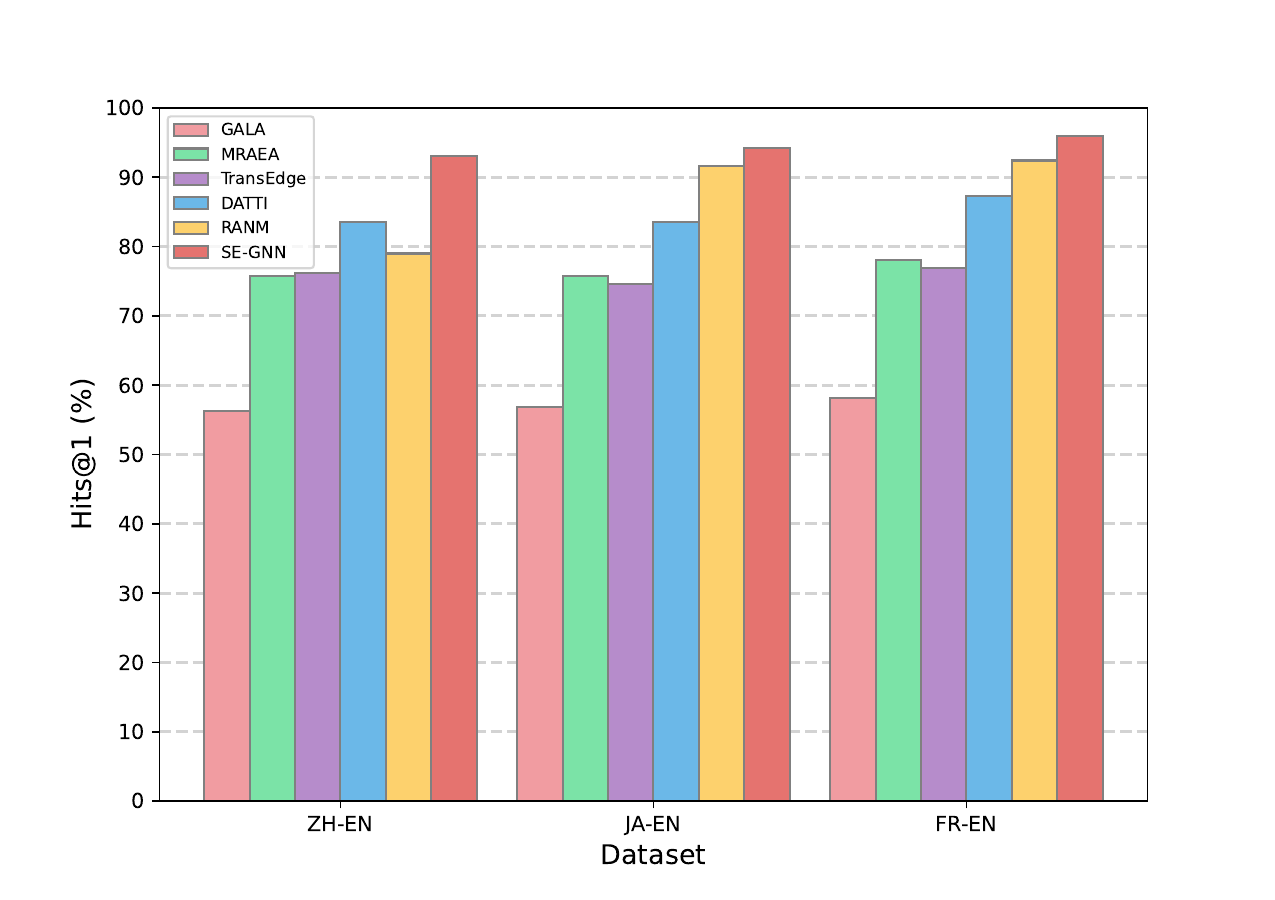}
		\caption{Compared with other semi-supervised models, the unsupervised SE-GNN performed using only neighborhood-level semantic information to construct a seed set has better results.}
		\label{unsuper}
	\end{figure}
	
	\subsection{\textbf{Unsupervised Entity Alignment Based on Embedding Model}}
	
	Based on the above research results, it is an effective method to supplement the alignment information with neighborhood-level entity semantic information and reduce the model's dependence on the seed set. Based on this conclusion, we tried an unsupervised entity alignment method without adding a pre-aligned seed set. We set the proportion of the seed set to 0\%, only use the neighborhood-level entity semantic information obtained by the BGE to construct the seed set, and then conduct experiments on the SE-GNN (semi) model on the DBP15K data set. As shown from Fig.\ref{unsuper}, SE-GNN still achieves excellent results without pre-aligned seed sets. This discovery provides us with a new entity alignment artifact. We first use the embedding model to obtain the semantic information representation of the entity, then build a seed set based on these representations by setting a screening strategy, and then bring it into the model for training. This unsupervised approach has the potential to achieve better results than supervised entity alignment and can lead to significant cost savings.

	\section{\textbf{Conclusions}}
	\label{section: section5}
	
	In this paper, we explore two problems the entity alignment task faces: existing methods mainly rely on single structural information to construct potential seed pairs and ignore the embedding distortion caused by noisy seed pairs in semi-supervised iterations. To this end, we propose SE-GNN. On the one hand, SE-GNN proposes a seed expansion strategy based on neighborhood-level semantic information, which comprehensively utilizes the semantic attributes and structural characteristics of entities to mine more and higher quality potential seed pairs fully. On the other hand, the threshold nearest neighbor embedding correction strategy of SE-GNN selects high-quality potential seed pairs and uses the embedding correction method to eliminate the embedding distortion caused by noisy seed pairs. Our overall experiments on the data set proved the superiority of SE-GNN, and at the same time, the ablation experiments also proved the importance of each module. At the same time, we also discovered the powerful effect of SE-GNN in using neighborhood-level entity semantic information to obtain rich alignment signals. However, SE-GNN still has room for improvement. SE-GNN needs to calculate many similarity matrices and aggregate high-order neighbor information, which means a lot of calculations. Our future work will focus on other strategies to supplement the alignment signal and reduce computational complexity while improving model performance.

	\bibliographystyle{IEEEtran}
	\bibliography{ref}

\end{document}